\documentclass[10pt,twocolumn]{article}

\usepackage[T1]{fontenc}
\usepackage[letterpaper,top=0.75in,bottom=1.0in,left=0.66in,right=0.66in,columnsep=0.33in]{geometry}
\usepackage{times}
\usepackage{graphicx}
\usepackage{caption}
\usepackage{booktabs}
\usepackage{algorithm}
\usepackage{algpseudocode}
\usepackage{amssymb}
\usepackage{amsmath}
\usepackage{longtable}
\usepackage{wrapfig}
\usepackage{comment}
\usepackage{soul}
\usepackage{xspace}
\usepackage{multirow}
\usepackage[table]{xcolor}
\usepackage{cite}
\usepackage[hidelinks]{hyperref}
\usepackage{cleveref}
\usepackage{microtype}

\definecolor{ourStudy}{RGB}{245, 194, 67}
\definecolor{compStudy}{RGB}{220, 133, 70}

\newcommand{\prag}[1]{{\noindent \textbf{#1}}}

\title{\vspace{-0.35in}\textbf{A Study of the Design Space of Motion and Disocclusion Control\\in Video Generation}\vspace{-0.08in}}

\author{%
Anran Qi$^{1}$ \quad
Changjian Li$^{2}$ \quad
Adrien Bousseau$^{1}$ \quad
Niloy J. Mitra$^{3,4}$\\[0.35em]
\small $^{1}$Centre Inria d'Université Côte d'Azur \quad
$^{2}$University of Edinburgh\\
\small $^{3}$Adobe Research \quad
$^{4}$UCL
}
\date{}

\begin{document}
\maketitle
\vspace{-0.25in}

\begin{figure*}[t]
  \centering
  \includegraphics[width=\textwidth]{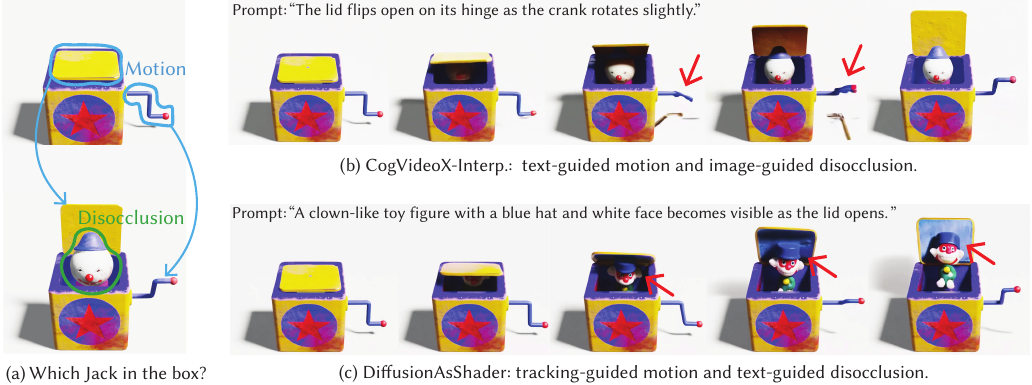}
  \caption{Every movement is a revelation. Like a Jack-in-the-Box slowly turning its crank, anticipation builds around the hidden spaces inside the box (a). Although object motion inherently creates disocclusions, motion control (\textit{what moves and how}) and disocclusion control (\textit{what appears in the newly revealed region}) remain largely unexplored as a unified design space. Methods emphasizing disocclusion control may generate unrealistic motion (b, video generated by CogVideoX-Interp.~\cite{cogvideoxinterp}), whereas methods emphasizing motion control may generate unexpected disocclusion content (c, video generated by DiffusionAsShader~\cite{diffAsShader}). In this work, we present a systematic study of this unified design space---so both the motion and the reveal are yours to script.}
  \label{fig:teaser}
\end{figure*}

\begin{abstract}
Disocclusion occurs when object movement reveals previously hidden content. Existing image-to-video methods provide different forms of motion and disocclusion control, yet the design space they span remains poorly understood. In this work, we systematically explore an algorithmic design space spanning motion specification (text-based versus tracking-based) and disocclusion specification (text-based versus image-based). To facilitate this study, we construct a benchmark consisting of both synthetic scenes and captured scenes, covering rigid and deformable motions, as well as static and dynamic disoccluded objects. We compare existing and adapted methods across this design space to analyze the strengths and limitations of different control strategies for handling disocclusions during motion. Our study reveals that tracking-based guidance improves motion accuracy, while image-based guidance improves the fidelity of newly revealed content. However, jointly controlling the appearance and dynamics of disoccluded objects remains a challenging open problem, particularly in scenarios where newly revealed objects undergo their own motion after becoming visible.
\end{abstract}

\noindent\textbf{Keywords:} Image-to-video generation, motion control, disocclusion

\section{Introduction}
\label{sec:intro}

Motion does not just displace—it reveals. As objects move, they uncover content that was previously hidden from view. While recent image-to-video generation methods provide increasingly explicit motion control through signals such as trajectories, optical flow, and dense tracking videos~\cite{geng2025motion, niu2024mofa, Wang2024Boximator, huang2025fine, phung2026trace, shi2024motion, diffAsShader, lee2026generative}, the appearance of newly revealed regions is typically inferred from the diffusion model's priors~\cite{ni2023conditional, brooks2024video, kong2024hunyuanvideo, blattmann2023stable}. We refer to the ability to control this revealed content as \emph{disocclusion control}. Image-conditioned interpolation methods partially address this problem by providing a target appearance reference~\cite{huang2022real, li2023amt, danier2024ldmvfi, jain2024video, xing2024dynamicrafter, feng2024explorative, wang2025framer}, but they generally offer weaker motion control than dedicated motion-control methods.

To systematically evaluate motion control and disocclusion control in image-to-video generation, we first construct a video benchmark, \textbf{VideoReveal}, consisting of both synthetic and captured scenes. The benchmark covers rigid and deformable object motions, static and dynamic disoccluded objects, and realistic appearance effects such as shadows and shading. The synthetic scenes provide accurate motion trajectories and disocclusion annotations, enabling controlled comparisons across methods. The captured scenes approximate practical editing workflows in which the first and last frames are available and the tracking video is recovered from sparse user input.

Equipped with this benchmark, we study the motion control and disocclusion control as two complementary dimensions of controllable video generation. For each, we compare weak, open-ended guidance (text prompt) with strong, explicit guidance (tracking video for motion and reference image for disocclusion). Their combination defines a simple $2 \times 2$ design space that enables a systematic comparison of motion and disocclusion control methods.

Mapping existing methods onto this design space reveals a missing quadrant: no prior approach combines explicit motion guidance with explicit appearance guidance. To fill this gap, we introduce a training-free extension of DiffusionAsShader~\cite{diffAsShader}, a tracking-based method that provides explicit motion control through tracking-video guidance. We augment it with reference-image conditioning inspired by interpolation-based video generation methods~\cite{cogvideoxinterp}. With all four quadrants instantiated, we can systematically compare different forms of motion and disocclusion specification, revealing their respective strengths, limitations, and the open challenges of jointly controlling motion and disocclusions.

Our study reveals distinct strengths of different control modalities. Tracking-based methods improve motion accuracy, while image-based guidance improves the fidelity of newly revealed content. However, jointly controlling the appearance and subsequent motion of disoccluded objects remains challenging, particularly when newly revealed objects must themselves move, deform, or interact with the scene. These findings highlight limitations of current image-to-video generation methods and suggest promising directions for future controllable video generation models.

In summary, our contributions are:

\begin{itemize}
    \item We introduce \emph{disocclusion control} as a complementary dimension to motion control in controllable image-to-video generation.

    \item We present a video benchmark comprising 32 synthetic cases and 5 captured scenes, enabling systematic evaluation of motion and disocclusion control across diverse scenarios.

    \item We formulate a two-dimensional design space spanning motion and disocclusion specification strategies, and instantiate all four quadrants using existing and adapted methods.

    \item We perform a comprehensive analysis of the resulting design space, revealing the strengths, limitations, and open challenges of current disocclusion-aware video generation methods.
\end{itemize}

\section{Related Work}
\label{sec:relatedWks}
\label{sec:Benchmark}
\begin{figure*}[ht]
    \centering
    \includegraphics[width=0.95\textwidth]{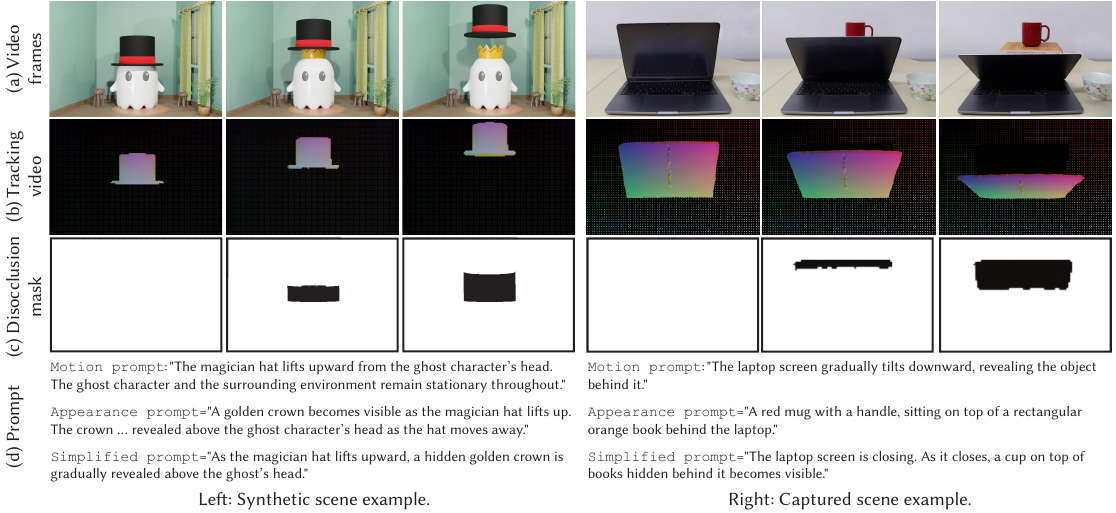}
    \vspace{-3mm}
    \caption{
    \textbf{Example synthetic and captured scenes from our benchmark.} (a) Rendered sequence in which the hat moves upward and reveals a hidden crown. The scene includes realistic illumination effects, such as the shadow cast by the hat over the wall and the highlight over the crown.
    (b) Tracking video used for motion control~\cite{diffAsShader}. (c) Disocclusion mask of the corresponding frames. (d) Associated prompts.
}

    \label{fig:dataset}
\end{figure*}

\paragraph*{Motion Control in Image-to-Video Generation.} Recent image-to-video (I2V) models enable animating a single image through various forms of motion guidance. Early approaches rely on text descriptions to specify the object motion \cite{xing2024dynamicrafter, xing2025motioncanvas, yang2024cogvideox,oh2018learning, chen2024videocrafter2, chen2023videocrafter1}. While intuitive, text often lacks the precision needed to describe exact object trajectories and dynamics, resulting in limited motion controllability. More recent methods provide stronger motion control by conditioning generation on motion signals, including sparse point trajectories \cite{wu2024draganything, Li2024PuppetMaster,geng2025motion, niu2024mofa,mou2024revideo}, bounding boxes \cite{Wang2024Boximator, huang2025fine, phung2026trace}, optical flows \cite{shi2024motion}, and dense tracking videos \cite{diffAsShader, physctrl2025, lee2026generative}. These representations offer increasingly precise control over what moves and how it moves, improving motion fidelity and user controllability. We choose DiffusionAsShader~\cite{diffAsShader} (abbreviated as DaS in later text) in our study as a representative motion control method, since its dense tracking-video conditioning provides a near-upper-bound reference for motion control, allowing us to isolate the effects of different disocclusion control strategies. Despite its strong controllability, tracking-based motion specifications often require users to provide detailed motion annotations during inference, which can limit practical usability~\cite{phung2026trace}. To study this trade-off, our benchmark includes both synthetic scenes, where accurate object trajectories are directly available from the underlying scene models, and captured scenes, where motion guidance is recovered from lightweight user interactions. The latter allows us to evaluate how tracking-based methods perform under more realistic usage scenarios.

Despite these advances in motion controllability, the appearance of disoccluded content is typically inferred by the generative model and remains weakly constrained~\cite{ni2023conditional, brooks2024video, blattmann2023stable}. In this work, we formulate motion control and disocclusion control as two orthogonal dimensions of controllable image-to-video generation and systematically study the design space they span.

\paragraph*{Disocclusion Control in Video Generation.} 
Disocclusion occurs when object motion reveals previously occluded regions that are not visible in the input image. Most image-to-video generation methods handle these newly revealed regions implicitly, relying on the generative model to hallucinate plausible content conditioned on the input image and text prompt~\cite{ni2023conditional, brooks2024video, kong2024hunyuanvideo, blattmann2023stable}. 

An alternative strategy is to control disocclusion through image-based appearance references. Video frame interpolation (VFI) is a representative example: intermediate frames are synthesized between given start and end frames, with the end frame providing an explicit appearance constraint.  Traditional VFI methods estimate optical flow between the input frames and synthesize intermediate frames through warping or splatting operations~\cite{huang2022real, li2023amt, zhang2023extracting}, but often struggle with large motions, complex occlusions, and long-range temporal dependencies~\cite{liu2024video}. More recently, diffusion-based approaches have reformulated VFI as a conditional video generation problem~\cite{danier2024ldmvfi, jain2024video, xing2024dynamicrafter, feng2024explorative,wang2025generative}. By leveraging large-scale video diffusion models, methods such as LDMVFI~\cite{danier2024ldmvfi} and VIDIM~\cite{jain2024video} can synthesize high-quality intermediate frames and better handle nonlinear motions. Wang et al.~\cite{wang2025framer}  further introduced interactive controls through user-specified sparse point trajectories during the interpolation. However, because only a small number of trajectories can be specified, it provides weaker motion control than dense tracking-based approaches that constrain the motion of every pixel of the entire object. This observation reveals a missing quadrant in the design space: no existing method combines explicit appearance guidance with explicit strong motion guidance. To fill this gap, we introduce a training-free variant.

\section{Benchmark}
\begin{figure*}[ht]
    \centering
    \includegraphics[width=1.0\textwidth]{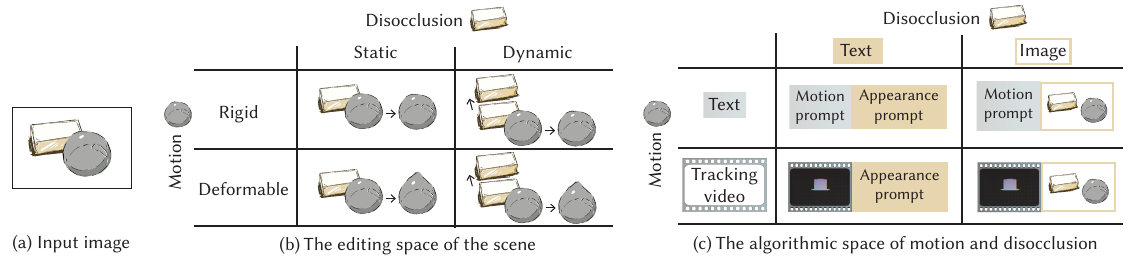}
    \vspace{-3mm}
    \caption{
    \textbf{Overview of the editing and algorithmic design spaces of disocclusion-aware image-to-video generation.}
    Starting from an input image (a), we consider four editing conditions formed by the combination of rigid/deformable motions and static/dynamic disoccluded objects (b). To perform such edits, existing algorithms vary in two dimensions (c): motion control (text prompt or tracking video) and disocclusion control (text prompt or reference image).
}
    \vspace{-3mm}
    \label{fig:space}
\end{figure*}

We introduce \textbf{VideoReveal}, a benchmark for systematic study of motion and disocclusion control in image-to-video generation. Each benchmark sample contains (1) \textbf{a ground-truth video} showing one or more objects moving within a scene and revealing previously occluded content. These videos serve as ground truth for evaluating image-to-video generation methods with varying forms of motion and disocclusion control (see Sec.~\ref{design_space}). 
(2) \textbf{A tracking video} (Fig.~\ref{fig:dataset} (b)) that encodes object motion as the motion guidance and is used as an explicit motion-control signal. Following \cite{diffAsShader}, the tracking video is rendered from dense point tracks sampled on the visible foreground regions of the first frame. Each point is assigned a color based on its first-frame position, with the reciprocal depth used as the $z$ coordinate. The colors remain fixed throughout the sequence, such that rendering the colored points at each timestep produces a video that encodes object motion while preserving point correspondences across frames. Since the tracking video is derived only from pixels visible in the first frame, it does not provide motion information for content that becomes visible later through disocclusion. To provide scene context, the static background is rendered sparsely as well. 
(3) \textbf{A per-frame disocclusion mask} (Fig.~\ref{fig:dataset} (c)) used to evaluate the disocclusion quality. A pixel is marked as disoccluded at frame $t$ onward if it is occluded in frame $t-1$ and becomes visible in frame $t$. (4) \textbf{Textual annotations} describing motion and disocclusion (Fig.~\ref{fig:dataset} (d)). Each sample is associated with a \emph{motion prompt} ($T_m$) describing the foreground motion, an \emph{appearance prompt} ($T_a$) describing the content revealed during disocclusion, and a \emph{simplified prompt} ($T_s$) providing a concise description of the overall scene dynamics and revealed content. These prompts are generated using GPT-5.5 \cite{chatgpt55} by providing the ground-truth video as input. The prompt-generation procedure is detailed in the supplementary material.

We construct the benchmark in two complementary settings. In the synthetic scenes, complete scene geometry and object motion are available, allowing us to generate accurate tracking videos and disocclusion masks while precisely controlling both motion and disocclusion events. This setting enables a systematic analysis of different motion and disocclusion control modalities under well-controlled conditions. In contrast, the captured scenes are designed to mimic realistic user workflows, where only the first and last images are available and object motion is recovered from lightweight user interactions rather than exact scene models. All benchmark videos have $49$ frames at a resolution of $720 \times 480$. We adopt these settings to match the generation configuration of CogVideoX-5B-I2V \cite{yang2024cogvideox}, the foundation model underlying all methods considered in our study. Together, these two settings allow us to evaluate both the controllability and practical usability of existing image-to-video generation approaches.

\subsection{Synthetic Scenes}
Figure~\ref{fig:dataset} shows an example synthetic scene in our benchmark.
We select 8 base scenes (Fig.~\ref{fig:dataset} (a)) spanning diverse environments and visual styles, including indoor and outdoor settings, daytime and nighttime scenes, and both photorealistic and stylized content. Across the benchmark, the moving objects cover both rigid and deformable motions, while the disoccluded objects can be either static or dynamic (Fig.~\ref{fig:space} (b)). For each scene, we design 2 distinct object motions, and for each motion, 2 disocclusion events that reveal different objects, resulting in 4 motion–disocclusion combinations per scene and 32 cases in total. To reflect real-world image formation, we incorporate scene lighting that produces shadows and shading effects on both moving and disoccluded objects, introducing realistic appearance changes during disocclusion events. We model all scenes in Blender, enabling the extraction of accurate object trajectories and disocclusion masks, which serve as precise motion and disocclusion control signals for generation and evaluation.

\subsection{Captured Scenes}
The captured scenes are designed to approximate realistic user workflows. Users only provide the first and last frame in which the desired disoccluded content is revealed. Unlike the synthetic scenes, no exact scene model or object trajectories are available. Therefore, we built a lightweight annotation tool to recover motion control signals from sparse user inputs. 

\noindent\textbf{Annotation Tool.} Given the first frame, we first employ MoGe~\cite{wang2025moge} to estimate the depth map and camera parameters (extrinsic and intrinsic). Next, we use SAM2~\cite{ravi2024sam} to segment objects or object parts through simple user interactions in the form of 2D bounding boxes. The resulting segments and camera parameters allow us to lift the scene into a set of point clouds. We then specify motion for movable objects by assigning motion parameters: (1) a motion type (translation or rotation), (2) a rotation or translation center selected by clicking a point in the input image, (3) a 3D motion direction specified through a disk-to-hemisphere mapping, where a 2D cursor position $(x,y)$ inside a unit circle is lifted to a unit hemisphere using $z=\sqrt{1-x^2-y^2}$, and (4) a transformation magnitude that we adjust to match the target pose in the final frame. The motion is linearly interpolated across the sequence, resulting in approximately constant-speed motion. These annotations are subsequently converted into tracking videos, used as motion guidance for image-to-video generation. A video demonstration of the tool is provided in the supplementary material.

The captured scenes focus on rigid-motion scenes to match the annotation tool's capacity.  We record 5 scenes using a Canon EOS 70D camera and capture the complete motion process from the initial to the final state. Each sequence is uniformly sampled to 49 frames and used as ground truth for quantitative evaluation. To focus the evaluation on motion and disocclusion control, we remove the hands used to move the objects during recording using an object-removal method~\cite{miao2026rose}, avoiding complex hand-motion interactions while preserving the object motion and disocclusion events.

\subsection{Evaluation Metrics}
As part of VideoReveal, we define the following evaluation metrics to assess four complementary aspects of the generated video: 

\noindent(a)~\textit{Motion Accuracy.}
We evaluate motion consistency at both the pixel and object levels. At the pixel level, we compare the optical flow fields of the generated and ground-truth videos. Specifically, we compute dense Farnebäck optical flow between consecutive frames in both videos. Motion accuracy (denoted as OptFlow) is measured as the mean cosine similarity between the resulting flow vectors at corresponding frame pairs throughout the sequence. At the object level, we use SAM2~\cite{ravi2024sam} to segment the moving object in both the generated and ground-truth videos. For each case, we identify the moving-object region from the first frame using the tracking video. We compute the Intersection-over-Union (denoted as IoU) between the generated and ground-truth masks for each frame and average it over the video. A higher IoU indicates better agreement with the target object motion and spatial extent.

\noindent(b)~\textit{Disocclusion Consistency.}
To evaluate whether newly revealed content matches the target appearance, we compare the generated video against the ground-truth video using two pixel-space metrics. First, we compute the average per-pixel Euclidean distance in RGB space over the entire frame sequence (denoted as Idiff). Second, we restrict the computation to the disoccluded regions specified by the disocclusion masks, yielding a masked variant (denoted as Idiff$_m$). Lower values indicate higher consistency with the desired appearance in both the overall scene and the newly revealed regions.

\noindent(c)~\textit{Semantic Similarity.}
Pixel-wise metrics may not fully capture perceptual consistency. Therefore, we additionally extract DINOv2~\cite{oquab2023dinov2} features from corresponding frames of the generated and ground-truth videos and compute their cosine similarity (denoted as DINO). Higher values indicate stronger semantic agreement between the generated content and the target video.

\noindent(d)~\textit{Video Quality.}
We evaluate video quality using FVDS~\cite{unterthiner2018towards,skorokhodov2022stylegan}, FVDC~\cite{unterthiner2018towards,yan2021videogpt}, SSIM~\cite{wang2004image}, PSNR~\cite{wang2004image}, and LPIPS~\cite{zhang2018unreasonable}. Intuitively, SSIM and PSNR measure low-level reconstruction fidelity, LPIPS assesses perceptual similarity, while FVDS and FVDC evaluate the realism of the generated video distribution and temporal dynamics.

\section{Evaluation}
\begin{figure*}[th]
    \centering
    \includegraphics[width=0.995\textwidth]{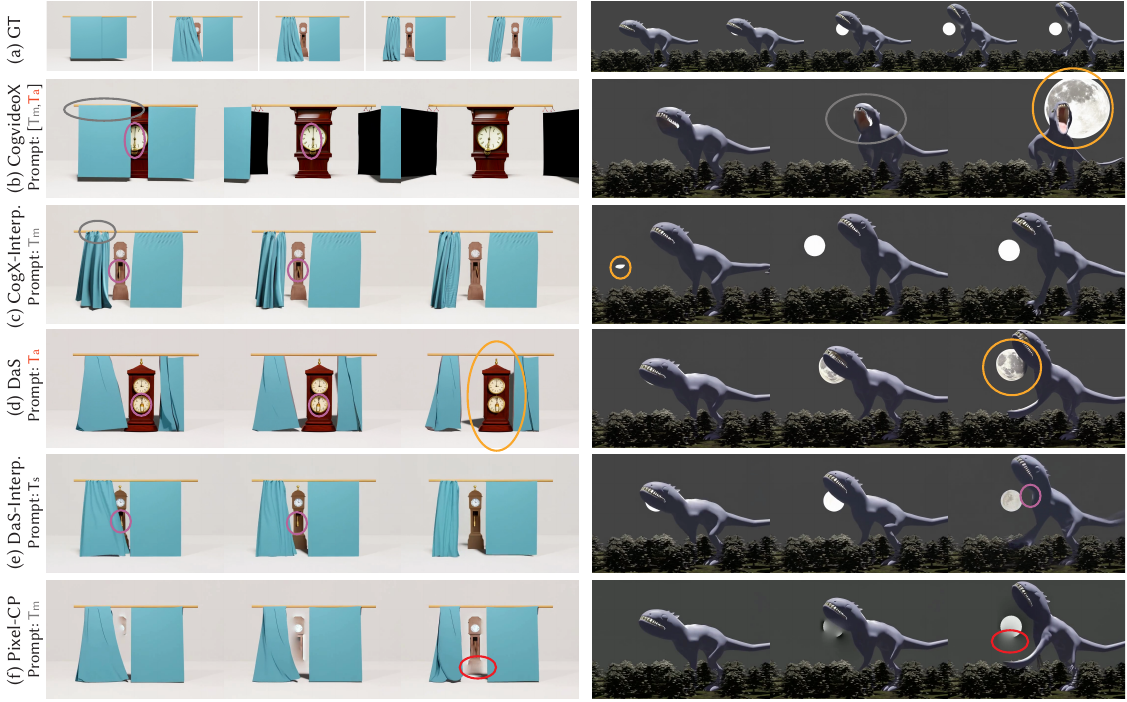}

    \caption{
    \textbf{Synthetic scene video results across the design space.} Grey circles highlight object motion, pink circles indicate shadowing and shading effects, orange circles mark the disoccluded objects, and red circles highlight visible artifacts. Both scenes involve deformable motion of the foreground object (the curtain and dinosaur body). The left scene reveals a dynamic disoccluded object (the swinging pendulum), whereas the right scene reveals a static one (the moon). The prompts used for the left example: 
\textcolor{gray}{\textbf{Motion prompt ($T_m$):}}
``The blue curtain on the left side slides smoothly ..." \textcolor{orange}{\textbf{Appearance prompt ($T_a$):}}
``A wooden grandfather clock becomes visible ... Inside the clock, a gold pendulum continuously swings left and right throughout the reveal." \textbf{Simplified prompt ($T_s$):}
``As the curtain and blue panel slide apart, a grandfather clock hidden behind them is gradually revealed, while the gold pendulum inside the clock swings side to side." 
}

\label{fig:video_model_results}
\end{figure*}

\subsection{CogVideoX Preliminaries}
In this work, we consider several methods built upon CogVideoX-5B-I2V~\cite{yang2024cogvideox}, either directly or through extensions such as DaS~\cite{diffAsShader}. We therefore briefly review the conditioning mechanism underlying CogVideoX before describing methods.

CogVideoX-5B-I2V (abbreviated as CogVideoX) is a latent diffusion model for image-to-video generation. Given an input image and a text prompt, the model synthesizes a video $\mathbf{V}\in\mathbb{R}^{49\times480\times720\times3}$ by iteratively denoising a latent video representation. The input image is encoded by the VAE into an image latent,  then it is concatenated with zero latents along the temporal dimension to form an image-conditioning tensor $\mathbf{Z}_c$ with the same temporal length as the video latent. During denoising, CogVideoX concatenates the noisy video latent and $\mathbf{Z}_c$ along the channel dimension and feeds the result to the transformer, while the text prompt is injected through text embeddings. After iterative denoising, the resulting latent is decoded by the VAE decoder $\mathcal{D}$ to produce the output video $\mathbf{V}$.

\subsection{Design Space of Motion and Disocclusion Control}
\label{design_space}

We study controllable image-to-video generation through two orthogonal dimensions: \textbf{motion control} and \textbf{disocclusion control}. Motion control specifies what moves and how it moves, whereas disocclusion control specifies the appearance of newly revealed regions. For motion control, we consider \emph{text-based motion}, which describes scene dynamics using natural language, and \emph{tracking-based motion}, which provides explicit object trajectories through a tracking video. For disocclusion control, we consider \emph{text-based disocclusion}, which specifies the revealed content through language, and \emph{image-based disocclusion}, which provides a reference image of the desired appearance. These choices define a $2\times2$ design space shown in Fig.~\ref{fig:space}(c). We instantiate this space using five representative methods. Tab.~\ref{tab:method_inputs} summarizes the inputs provided to each method. We note that motion control and disocclusion control are two orthogonal aspects of controllable video generation. Yet, the conditioning modalities (text vs. tracking video for motion, and text vs. reference image for disocclusion) are not mutually exclusive. The goal of our study is to analyze the strengths and limitations of each conditioning modality within a unified design space, rather than to suggest that these modalities cannot be combined.

\begin{table}[t]
\centering
\caption{Summary of the inputs provided to each evaluated method.
$T_m$, $T_a$, and $T_s$ denote the motion, appearance, and simplified
prompts, respectively.}
\label{tab:method_inputs}
\resizebox{0.95\columnwidth}{!}{
\begin{tabular}{lcccc}
\toprule
Method & Initial Image & Text Prompt & Tracking Video & Reference Image \\
\midrule
CogVideoX    & \checkmark & $T_m + T_a$ & --         & -- \\
CogX-Interp. & \checkmark & $T_m$       & --         & \checkmark \\
DaS          & \checkmark & $T_a$       & \checkmark & -- \\
DaS-Interp.  & \checkmark & $T_s$       & \checkmark & \checkmark \\
Pixel-CP     & \checkmark & $T_m$       & \checkmark & \checkmark \\
\bottomrule
\end{tabular}}
\end{table}

\noindent\textbf{1. CogVideoX}~\cite{yang2024cogvideox} occupies the text-motion/text-disocclusion quadrant, where both motion and newly revealed content are specified through language (\emph{motion prompt} and \emph{appearance prompt}). 

\noindent\textbf{2. CogVideoX-Interpolation}~\cite{cogvideoxinterp} (abbreviated as CogX-Interp.) occupies the text-motion/image-disocclusion quadrant. It extends CogVideoX by using a \emph{motion prompt} to describe motion and the last frame to specify the disocclusion content. Specifically, the model is fine-tuned to perform keyframe interpolation by augmenting the image-conditioning tensor $\mathbf{Z}_c$ with the encoded first-frame feature, zero-padded intermediate features, and the encoded target last-frame feature. This allows the model to generate intermediate frames that transition from the input image to the target frame while preserving the desired disoccluded appearance.
 
\noindent\textbf{3. DaS}~\cite{diffAsShader} occupies the tracking-motion/text-disocclusion quadrant. DaS is a fine-tuned variant of CogVideoX that introduces a tracking video to explicitly encode object trajectories over time and uses an \emph{appearance prompt} for disocclusion. The tracking video is encoded using the same VAE as the input image and serves as an additional conditioning signal during denoising. By replacing text-based motion descriptions with explicit trajectory guidance, DaS enables more precise motion control.
 
\noindent The fourth quadrant combines dense tracking-based motion control with image-based disocclusion specification. To the best of our knowledge, no existing method occupies this quadrant. We therefore introduce two instantiations: \noindent\textbf{4. DaS-Interp.} is a training-free extension of DaS which combines tracking-video guidance with image-based disocclusion control. Specifically, we retain DaS's tracking-video conditioning for motion control while incorporating the same target-frame conditioning strategy used in CogX-Interp. by providing the last frame as an appearance constraint. This formulation enables the model to simultaneously follow the prescribed motion trajectories and match the target disoccluded appearance. Since both motion and disocclusion are specified, we use the \emph{simplified prompt}. See Supplementary Material for implementation details. \textbf{5. Pixel-CP.} Unlike DaS-Interp., which incorporates the target appearance during video generation, Pixel-CP enforces appearance control through post-processing. It first uses DaS to generate a video guided by the tracking video, complemented by the \emph{motion prompt} to describe the desired motion without information about the appearance of the disoccluded object. The disoccluded regions are then directly copied from the reference image and composited into the generated frames using the disocclusion masks and Poisson image editing. Its performance is therefore affected by motion-generation errors, inaccuracies in the disocclusion masks, and blending artifacts. Moreover, direct compositing cannot reproduce global appearance effects induced by the revealed content, such as shadows or reflections.

 Together, these methods instantiate all quadrants of the algorithmic design space, enabling a systematic comparison of motion and disocclusion control approaches. Implementation details are provided in Suppl.~Secs.~2 and 3.

\begin{table*}[t]
\centering
\scriptsize
\setlength{\tabcolsep}{2.8pt}
\renewcommand{\arraystretch}{1.12}

\caption{\textbf{Quantitative results for image$\rightarrow$video across the design space.}
Two settings: \emph{Synthetic scenes} (top) and \emph{Captured scenes} (bottom). The \emph{Alg. Space} column identifies the algorithmic design-space category of each method.
Higher is better for OptFlow/IoU/SSIM/PSNR; lower is better for Idiff/Idiff$_m$/FVD-S/FVD-C/LPIPS. 
Best and second-best are \textbf{bold} and \underline{underlined}.  }

\label{tab:eval_comparison}
\resizebox{0.975\textwidth}{!}{%
\begin{tabular}{llcccccccccc}
\toprule
\textbf{Method} & \textbf{Alg. Space} & \textbf{OptFlow $\uparrow$}  & \textbf{IoU $\uparrow$} & \textbf{Idiff $\downarrow$} & \textbf{Idiff$_m$ $\downarrow$} & \textbf{DINO $\uparrow$} & \textbf{FVD-S $\downarrow$} & \textbf{FVD-C $\downarrow$} & \textbf{SSIM $\uparrow$} & \textbf{PSNR $\uparrow$} & \textbf{LPIPS $\downarrow$} \\
\midrule
\multicolumn{12}{c}{\cellcolor{gray!12}\textbf{Synthetic scenes}} \\
CogVideoX & Text-Text & -0.006 & 0.351& 74.41 & 34.07 & 0.733 & 1580.7 & 1584.7 & 0.828 & 17.71 & 0.267 \\ 
\arrayrulecolor{black!10}
\hline

CogX-Interp. &Text-Image  & 0.208 &0.536& 42.99 & 20.81 & \underline{0.841} & 960.6 & 963.0 & 0.873 & 20.45 & 0.182 \\ \hline
DaS & Track-Text & \underline{0.681} &0.728 & 45.22 & 19.46 & 0.801 & 878.7 & 881.1 & 0.881 & 21.45 & 0.146 \\\hline
DaS-Interp. &  \multirow{2}{*}{Track-Image} & 0.618 & \textbf{0.778}& \textbf{32.16} & \textbf{17.70} & 0.837 & \underline{709.4} & \underline{711.5} & \textbf{0.903} & \textbf{23.60} & \textbf{0.125} \\
Pixel-CP &  & \textbf{0.694} &\underline{0.772}& \underline{35.56} & \underline{17.96} & \textbf{0.862} & \textbf{653.0} & \textbf{654.7} & \underline{0.889} & \underline{22.76} & \underline{0.132} \\
\addlinespace[2pt]
\arrayrulecolor{black}

\midrule
\multicolumn{12}{c}{\cellcolor{gray!12}\textbf{Captured scenes}} \\
CogVideoX & Text-Text & 0.037 &0.337& 144.19 & 70.86 & 0.716 & 2358.1 & 2364.2 & 0.657 & 13.13 & 0.453 \\ \arrayrulecolor{black!10}  \hline
CogX-Interp. & Text-Image & 0.701 &0.851& \textbf{37.65} & \textbf{17.69} & \textbf{0.961} & \textbf{381.0} & \textbf{382.4} & \textbf{0.845} & \textbf{22.68} & \textbf{0.124} \\ \hline
DaS & Track-Text & \underline{0.741} & 0.796& 67.49 & 25.15 & 0.880 & 697.4 & 699.7 & 0.797 & 20.23 & 0.187 \\\hline
DaS-Interp. & \multirow{2}{*}{Track-Image} & 0.711 &\textbf{0.869}& \underline{47.06} & \underline{22.08} & 0.880 & 862.8 & 865.3 & \underline{0.830} & \underline{21.15} & 0.174 \\
Pixel-CP &  & \textbf{0.789} & \underline{0.860}& 56.48 & 25.36 & \underline{0.904} & \underline{611.3} & \underline{613.0} & 0.794 & 20.54 & \underline{0.172} \\
\arrayrulecolor{black}
\bottomrule
\end{tabular}%
}
    \vspace{-2mm}
\end{table*}

\begin{figure}[!h]

    \centering
    \includegraphics[width=1.0\columnwidth]{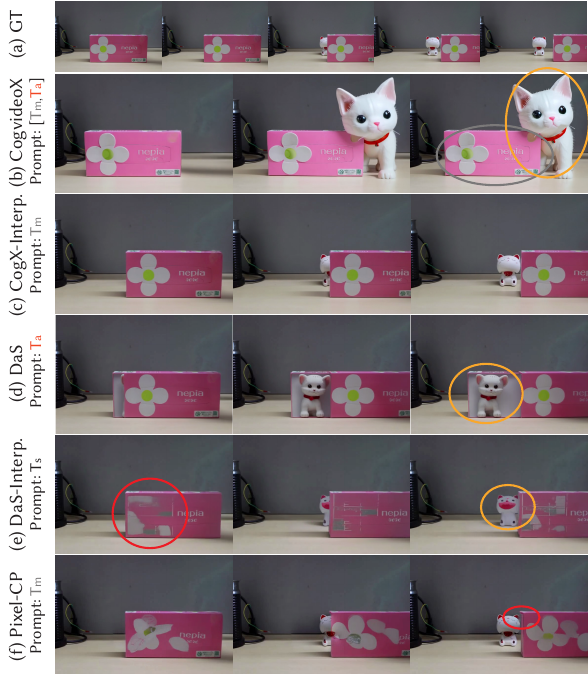}
    \vspace{-3mm}
    \caption{
    \textbf{Captured scene results.} This scene combines rigid foreground motion (tissue box slides away) with a dynamic disoccluded object (the cat continuously nods its head up and down). Gray circles highlight object motion, orange circles mark the disoccluded objects, and red circles highlight visible artifacts. 
\textcolor{gray}{\textbf{Motion prompt ($T_m$):}}
``The pink tissue box slides smoothly toward the right across the tabletop." \textcolor{orange}{\textbf{Appearance prompt ($T_a$):}}
``A small white cat ... becomes visible as the pink tissue box slides ... the cat continuously nods its head up and down."  \textbf{Simplified prompt ($T_s$):}
``As the tissue box slides ... the cat continuously nods its head up and down." }
        \vspace{-5mm}
        \label{fig:video_capture_results}
\end{figure}

\section{Results}
\label{sec:experiments}

\subsection{Visual Analysis}

\paragraph*{Synthetic Scenes.}
Fig.~\ref{fig:video_model_results} shows the synthetic scene results and example prompts used. We observe that first, generative methods produce realistic shading and shadow variations throughout the scene around newly revealed regions. For example, DaS and DaS-Interp. reproduce the shadow changes around the pendulum and shading around the dinosaur's neck (highlighted by pink circles), whereas Pixel-CP directly copies pixels from the reference image and therefore lacks these secondary effects. Second, settings using tracking videos (DaS, DaS-Interp., and Pixel-CP) produce more accurate object motion than text-only settings (highlighted by grey circles). Third, text-based disocclusion control often produces semantically plausible but visually different objects (orange circles). In the curtain example, CogVideoX and DaS reveal clock-like objects, but fail to reproduce the specific target clock. In contrast, image-conditioned methods preserve the desired object identity. While Pixel-CP reliably reproduces the correct disoccluded object because it directly copies content from the reference image, this strategy introduces visible copy-and-paste artifacts (red circles). 

In the curtain scene (left), all generative approaches correctly animate the pendulum, likely because pendulum-like oscillatory motion is well represented in the data used to train the underlying generative model. However, this capability does not generalize reliably. In the captured scene (Fig.~\ref{fig:video_capture_results}), the prompt specifies that ``the cat continuously nods its head up and down'', yet all methods generate a static head. This qualitative observation is consistent with the quantitative analysis in Sec.~\ref{sec:quantitative}, which shows that dynamic disocclusion remains more challenging than static disocclusion. Since neither tracking videos nor reference images provide explicit motion constraints for content that is initially occluded, the subsequent dynamics of disoccluded objects remain largely unconstrained.

\paragraph*{Captured Scenes.}
Fig.~\ref{fig:video_capture_results} shows captured-scene results. Compared with the synthetic benchmark, motion guidance is less reliable because tracking videos are estimated from sparse user annotations rather than derived from an exact scene model. As a result, object motion may become spatially incoherent—for example, only the box lid follows the prescribed motion in Fig.~\ref{fig:video_capture_results}(d)—and appearance artifacts such as flower-texture stretching and local deformations may emerge (Fig.~\ref{fig:video_capture_results}(e,f)). DaS-Interp. preserves the overall object structure but tends to lose fine details of the revealed object, producing a slightly funny Maneki-neko mouth (Fig.~\ref{fig:video_capture_results}(e), orange circle). In contrast, CogX-Interp. achieves near-perfect results for this example (Fig.~\ref{fig:video_capture_results}(b)). Since the scene contains relatively simple translational motion and the target appearance is explicitly provided by the reference image, the model can effectively interpolate  while preserving both object identity and visual quality. This result suggests that image-conditioned interpolation performs well for simple motions. However, errors in the estimated trajectories can degrade both motion accuracy and generation quality, demonstrating that tracking-based control strongly depends on the quality of the provided motion guidance. This highlights the need for more reliable motion-specification and tracking tools to provide accurate and coherent motion signals in practical settings.

\subsection{Quantitative Results Analysis}
\label{sec:quantitative}

\prag{Algorithmic Design Space.} The quantitative results are reported in Tab. \ref{tab:eval_comparison}. For motion accuracy, tracking-conditioned settings generally achieve higher motion accuracy than text-only settings. In the synthetic scenes, Pixel-CP achieves the highest OptFlow score, closely followed by DaS, while CogVideoX exhibits near-zero motion agreement. Similar trends are observed in the captured scenes. These results indicate that text prompts alone are insufficient for specifying precise object trajectories, whereas tracking-based guidance provides reliable motion control.

For disocclusion consistency, in the synthetic scenes, DaS-Interp. obtains the lowest Idiff and Idiff$_m$, while Pixel-CP achieves comparable results. In the captured scenes, CogX-Interp. achieves the best disocclusion scores, benefiting from the availability of the reference image, while DaS-Interp. and Pixel-CP remain competitive. Overall, image-conditioned settings achieve higher disocclusion fidelity than text-only settings, confirming that images provide stronger constraints on object appearance and geometry. The DINO scores confirm the importance of disocclusion specification for preserving semantic consistency. 

Regarding video quality, Track-Image methods again perform favorably in the synthetic scenes. DaS-Interp. achieves the best SSIM, PSNR, and LPIPS scores, while Pixel-CP obtains the lowest FVD values, indicating strong fidelity to the target video. In the captured scenes,  since the tracking is estimated from sparse user annotations rather than derived from a known scene model, tracking inaccuracies introduce an additional source of error (e.g., texture changes, incoherent object motion), which makes the tracking-based method less competitive. Instead, CogX-Interp. achieves the best scores across nearly all quality metrics. The results suggest that image-based disocclusion specification is consistently important across both controlled and practical settings, while the benefit of explicit motion guidance depends
on the quality of the available motion information.

To complement these metrics, we additionally conduct a perceptual user study based on pairwise preference comparisons. The study protocol and detailed results are provided in Suppl.~Sec.~1.

\prag{Editing space.} 
In Tab. \ref{tab:edit_comparison}, we analyze performance with respect to two editing attributes: foreground motion type (rigid vs. deformable) and disoccluded object behavior (static vs. dynamic). As the synthetic benchmark includes all four conditions and provides accurate motion tracking and disocclusion annotations, we use it for this analysis.

The OptFlow results highlight the impact of object deformation. Across nearly all methods, rigid motions yield higher flow consistency than deformable motions, confirming that non-rigid shape changes remain difficult to reproduce accurately. Nonetheless, methods with explicit motion tracking maintain relatively strong performance under both rigid and deformable motions.
For $\mathrm{Idiff}$, dynamic scenes are generally more challenging than static scenes, indicating that complex object motions increase appearance discrepancies in newly revealed regions.  The inset figure reports the motion generation success rate of the disoccluded objects for cases involving  dynamic
\setlength{\columnsep}{0.1in}
\begin{wrapfigure}[9]{r}{0.6\columnwidth}
    \vspace{-3mm}
    \includegraphics[width=\linewidth]{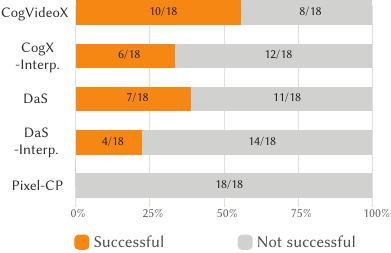}
    \label{fig:dynamic_disocclusion}
\end{wrapfigure}
 disoccluded objects (18 cases in total, including 16 synthetic scenes and 2 captured scenes). A case is considered successful if the generated motion exhibits the intended motion type (e.g., translation, rotation, or articulation) and follows the expected motion pattern throughout the sequence.

CogVideoX achieves the highest success rate, correctly reproducing the intended motion in 10 out of 18 cases, followed by DaS. Since DaS tracking videos only specify the motion of pixels visible in the first frame, the motion of newly revealed objects remains unconstrained. This suggests that additional motion layers covering disoccluded regions may be necessary to improve the motion control of disoccluded objects. Methods that additionally enforce explicit image-based disocclusion constraints (CogX-Interp. and DaS-Interp.) exhibit lower success rates.  This suggests that preserving consistency with a target appearance can interfere with the generation of correct object dynamics. Pixel-CP fails in all dynamic cases because it directly copies the disoccluded content from the target image and therefore cannot synthesize the required object motion. Overall, these results indicate that dynamic disocclusion remains challenging for current image-to-video generation models, as they must jointly reason about the appearance of newly revealed content and its subsequent motion.

\begin{table}[t]
\centering
\small
\caption{Comparison across editing properties in synthetic scenes. For Idiff, lower is better. For OptFlow cosine similarity, higher is better. Within each pair of conditions, the better result is shown in \textbf{bold}.}
\label{tab:edit_comparison}

\resizebox{0.95\columnwidth}{!}{%
\begin{tabular}{lcccc}
\toprule
& \multicolumn{2}{c}{\textbf{Idiff $\downarrow$}}
& \multicolumn{2}{c}{\textbf{OptFlow $\uparrow$}} \\
\cmidrule(lr){2-3}
\cmidrule(lr){4-5}
Method & Static & Dynamic & Rigid & Deformable \\
\midrule
CogVideoX
& \textbf{66.8} & 82.0
& -0.022 & \textbf{0.010} \\

CogX-Interp.
& 46.2 & \textbf{39.8}
& \textbf{0.213} & 0.203 \\

DaS
& \textbf{43.2} & 47.3
& \textbf{0.853} & 0.509 \\

DaS-Interp.
& \textbf{26.9} & 37.8
& \textbf{0.769} & 0.476 \\

Pixel-CP
& \textbf{28.0} & 43.1
& \textbf{0.847} & 0.541 \\
\bottomrule
\end{tabular}%
}
    \vspace{-2mm}
\end{table}

\section{Limitations and Future Work}

Our study reveals several open challenges for disocclusion-aware image-to-video generation.

First, dynamic disocclusion remains an open challenge for controllable video generation. Existing control mechanisms can specify the appearance of newly revealed content, but provide limited control over its subsequent motion. As a result, scenarios in which disoccluded objects must themselves move, deform, or interact with the scene remain particularly difficult. Future work could explore motion representations that explicitly model multiple layers of scene motion, allowing motion specifications to be attached to both visible and disoccluded objects.

Second, our instantiation of the Track-Image quadrant, DaS-Interp., is a training-free combination of existing motion and appearance control mechanisms. While this design enables a controlled comparison within the proposed design space, it is not optimized to 
satisfy both motion and disocclusion constraints. Future work could investigate lightweight adaptation strategies, such as adapters~\cite{T2I-Adapter, IP-Adapter} or LoRA modules~\cite{LoRA}, to better integrate motion and disocclusion control.

Third, accurate motion signal specification remains a key challenge for controllable video generation. In practical settings, tracking videos are often estimated from sparse user interactions, and inaccuracies in the recovered motion can lead to visible artifacts such as incoherent object motion. Future work could investigate intuitive interfaces and techniques that enable users to specify accurate object motion~\cite{doersch2022tap}. Extending these interfaces to support deformable motions  (e.g., ARAP or linear blend skinning (LBS) for rig-based deformation, or learned controllable blendshapes) and highly free-form dynamics (e.g., water splashes, fluttering flags, hair in wind) is an important future direction.

Fourth, our benchmark is limited in scale, comprising 32 synthetic cases and 5 captured scenes. This limits the extent to which our observations can be generalized to the diversity and complexity of real-world videos. In particular, most scenes involve a limited number of objects and localized disocclusion events, whereas real-world scenarios may contain multiple interacting objects or large-scale scene reveals. In addition, all methods evaluated in this study are based on the CogVideoX model family, either directly or through its extensions. Consequently, our findings should be interpreted as observations within the evaluated benchmark and model family, rather than as universal conclusions about video generation models. Extending VideoReveal to a larger and more diverse set of scenes and evaluating additional video-generation model families are important directions for future work.

Finally, this work focuses on disocclusion caused by object motion. In many real-world videos, newly visible content also arises from camera motion, including panning, zooming, and viewpoint changes~\cite{xing2025motioncanvas, he2024cameractrl,wang2025cinemaster}. Handling camera-induced disocclusions introduces additional challenges related to scene geometry and depth reasoning, and remains an open problem.

\section{Conclusion}

We presented a benchmark and systematic study of motion and disocclusion control in image-to-video generation through a two-dimensional design space. Across both visual and quantitative evaluations, we observe that tracking-conditioned settings generally achieve higher object-motion accuracy than text-only settings, while image-based disocclusion guidance substantially improves the fidelity and identity preservation of newly revealed content. 
In practical captured-scene settings, image-conditioned interpolation performs particularly well for simple motions, while tracking-based methods become sensitive to errors in the estimated trajectories. This sensitivity highlights the importance of developing better motion-specification tools that provide accurate and reliable motion guidance for video generation models. Additionally, dynamic disocclusion remains especially challenging because models must jointly reason about the appearance and subsequent motion of newly revealed objects. Overall, our results suggest that effective disocclusion-aware video generation requires both reliable motion specification and explicit appearance constraints. While current methods can often satisfy one of these objectives, jointly controlling the appearance and dynamics of newly revealed content remains a challenging open problem.

\section*{Acknowledgments}
We thank Zekai Gu for insightful discussions on Diffusion as Shader, Mia Kan for assistance with the user study, and Berend Baas for proofreading. This work was partially supported by JST ASPIRE JPMJAP2401, gifts from Adobe and UCL AI Centre. CL was supported by a gift from Adobe.

\bibliographystyle{unsrt}
\bibliography{bibpg}

\end{document}